\documentclass[journal]{IEEEtran}
\ifCLASSINFOpdf
\usepackage[pdftex]{graphicx}
\graphicspath{{figure/}}
\else
\fi
\usepackage{amsmath}

\usepackage{cite}
\usepackage{multirow}
\usepackage{booktabs} 
\usepackage{amssymb}
\usepackage{pifont}
\usepackage{makecell}
\newcommand{\cmark}{\ding{51}}%
\newcommand{\xmark}{\ding{55}}%

\begin{document}
%
\title{HAN: An Efficient Hierarchical Self-Attention Network for Skeleton-Based Gesture Recognition}
%
%
%

\author{Jianbo~Liu,
	Ying~Wang,
	Shiming~Xiang~\IEEEmembership{Member,~IEEE,}
	and~Chunhong~Pan,~\IEEEmembership{Member,~IEEE}
	\thanks{Corresponding author: Ying Wang.}
	\thanks{Jianbo Liu and Shiming Xiang are with the National Laboratory of Pattern Recognition, Institute of Automation, Chinese Academy of Sciences, Beijing 100190, China, and also with the School of Artificial Intelligence, University of Chinese Academy of Sciences, Beijing 101408, China (e-mail: jianbo.liu@nlpr.ia.ac.cn; smxiang@nlpr.ia.ac.cn).}
	\thanks{ Ying Wang and Chunhong Pan are with the National Laboratory of Pattern Recognition, Institute of Automation, Chinese Academy of Sciences, Beijing 100190, China (e-mail: ywang@nlpr.ia.ac.cn; chpan@nlpr.ia.ac.cn).}
}

%
%

\markboth{Journal of \LaTeX\ Class Files,~Vol.~14, No.~8, August~2015}%
{Shell \MakeLowercase{\textit{et al.}}: Bare Demo of IEEEtran.cls for IEEE Journals}
%



\maketitle

\begin{abstract}
	Previous methods for skeleton-based gesture recognition mostly arrange the skeleton sequence into a pseudo picture or spatial-temporal graph and apply deep Convolutional Neural Network (CNN) or Graph Convolutional Network (GCN) for feature extraction. Although achieving superior results, these methods have inherent limitations in dynamically capturing local features of interactive hand parts, and the computing efficiency still remains a serious issue. In this work, the self-attention mechanism is introduced to alleviate this problem.  Considering the hierarchical structure of hand joints, we propose an efficient hierarchical self-attention network (HAN) for skeleton-based gesture recognition, which is based on pure self-attention without any CNN, RNN or GCN operators. Specifically, the joint self-attention module is used to capture spatial features of fingers, the finger self-attention module is designed to aggregate features of the whole hand. In terms of temporal features, the temporal self-attention module is utilized to capture the temporal dynamics of the fingers and the entire hand. Finally, these features are fused by the fusion self-attention module for gesture classification. Experiments show that our method achieves competitive results on three gesture recognition datasets with much lower computational complexity.
\end{abstract}


%
\IEEEpeerreviewmaketitle

\section{Introduction}
%
%
%
%
\IEEEPARstart{G}{esture} recognition is a popular research topic in the field of computer vision with applications in many fields, such as human computer interaction and sign language interpretation. Previous works for gesture recognition mainly take RGB-D videos as input, then various convolutional neural networks (CNNs) are designed to capture the spatial and temporal features for hand gestures. Recently, with great progress made in hand pose estimation, accurate coordinates of hand joints can be easily obtained. Thus, skeleton has become a popular modality for action and gesture recognition. A variety of works concentrate on predicting gesture categories by sequences of hand joints with 3D coordinates. Compared with images, skeleton is more robust to varying background noise. Moreover, it is easier to design a lightweight model for gesture recognition owing to the small data size of the skeleton. Therefore, we focus on the skeleton-based method in this work.

\begin{figure}[!t]
	\centerline{\includegraphics[width=\columnwidth]{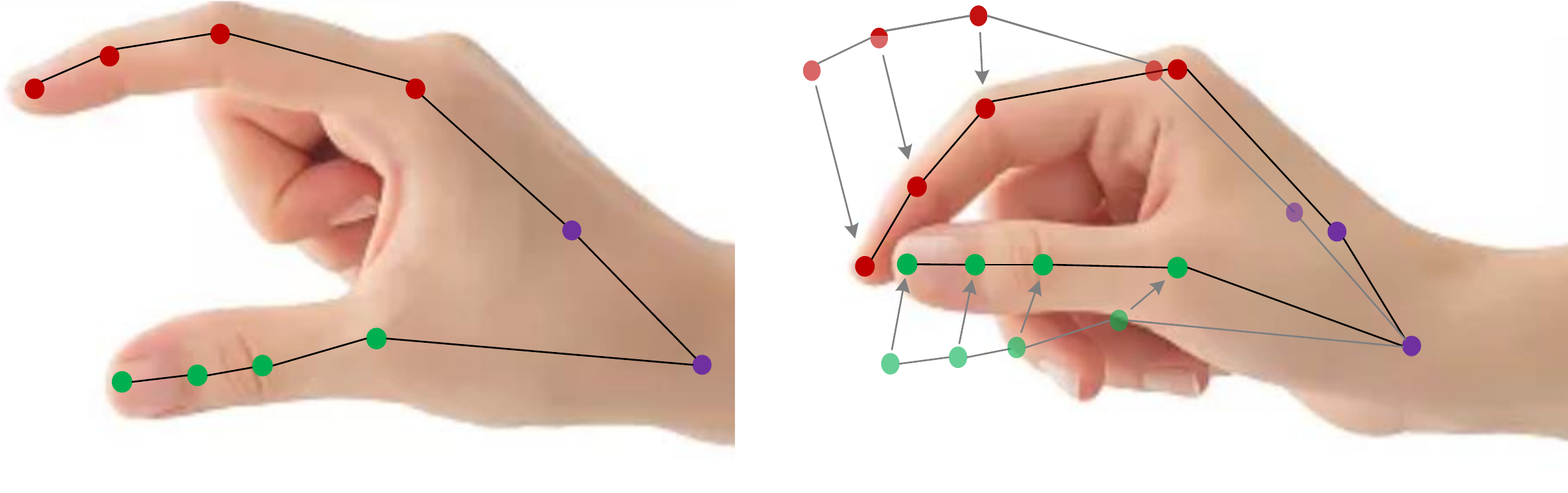}}
	\caption{The motion of hand joints for gesture ``Pinch". The arrows represent the movements of hand joints. The motion of four joints in each finger has a similar tendency.}
	\label{fig:pinch}
\end{figure}


Motivated by the great success of deep learning on computer vision tasks, recent studies~\cite{chen2017motion,hou2018spatial,nunez2018convolutional} aim to apply deep neural networks to skeleton-based gesture recognition. In these studies, some works arrange hand skeleton sequence as a pseudo image~\cite{nunez2018convolutional}, where frames are treated as the column of the image, hand joints are treated as the row, and the 3D coordinates are corresponding to the three channels of the image. Accordingly, 2D CNN based neural network can be used to extract spatial and temporal features for gesture or action recognition. More recently, a host of methods~\cite{yan2018spatial} embed a skeleton sequence into a spatial-temporal graph, where joints within each frame are connected by the underlying structure of body skeleton, and the same joints in adjacent frames are connected as well. Then graph convolutional networks are devised to capture discriminating cues for action recognition. Although these methods have achieved excellent performance, they have inherent limitations in capturing local features of interactive joints. As the skeleton sequence is embedded into a predefined image or graph with a fixed structure, the interactive joints may not be adjacent to each other. Therefore, only the deep layers can aggregate the information of these interactive joints. Liu {\em et al.} propose to model the skeleton sequence into a 3D volume, a 3D CNN based network is utilized to extract skeleton dynamics~\cite{liu2020decoupled}. By volume modeling for skeletons, the drawbacks described above are alleviated. However, the 3D CNN based network is computationally expensive to be deployed on a device with limited computing resources.

\begin{figure*}[!t]
	\centerline{\includegraphics[width=\textwidth]{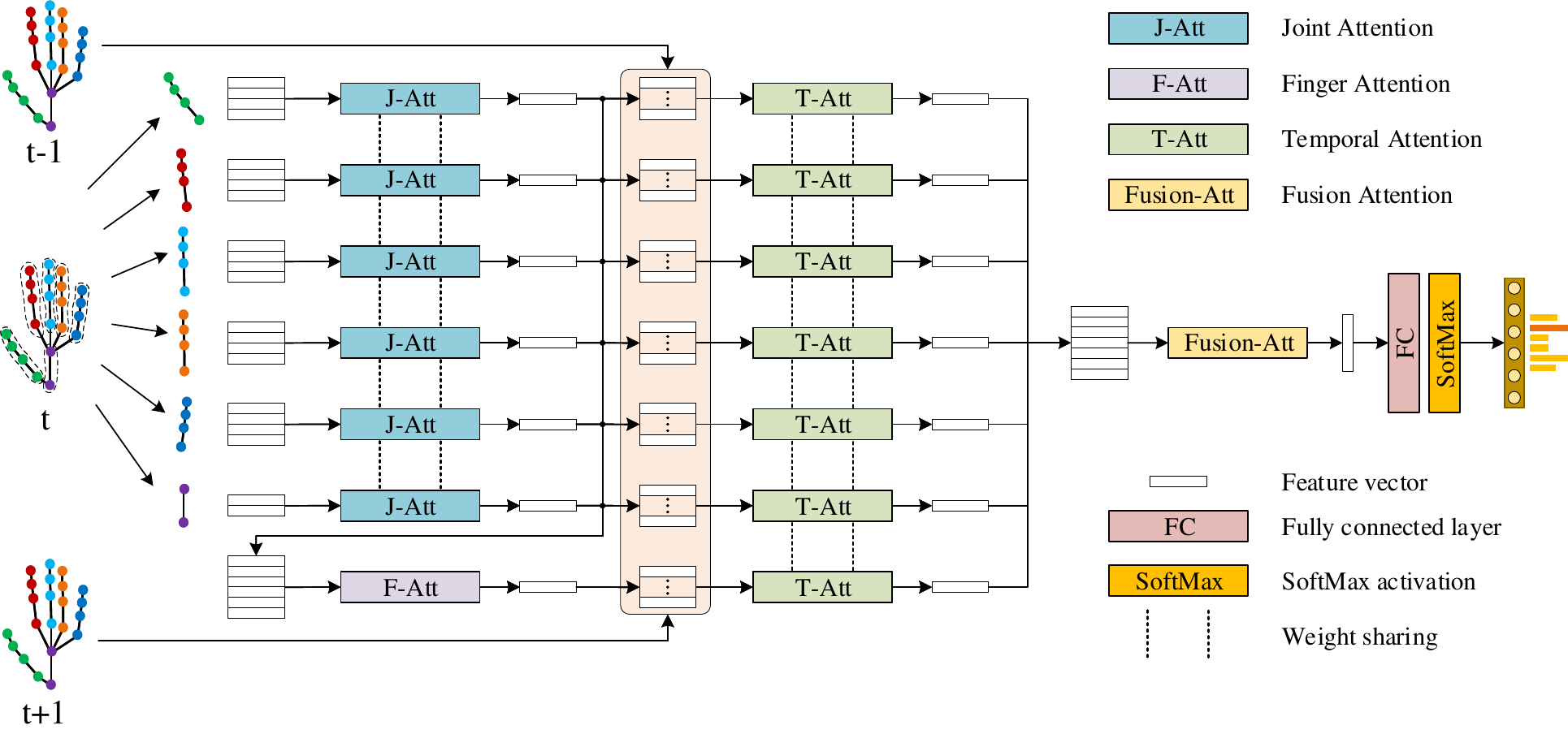}}
	\caption{The framework of our proposed Hierarchical Self-Attention Network (HAN). The hand is divided into 6 parts, joints of each part are fed into \mbox{J-Att} to extract finger features. These finger features are aggregated to hand features using \mbox{F-Att}. \mbox{T-Att} is used to capture temporal features of fingers and hand. Finally, \mbox{Fusion-Att} is utilized to fuse the temporal features for gesture recognition. \mbox{J-Att} and \mbox{T-Att} share weights with different inputs.}
	\label{fig:framework}
\end{figure*}

On the other hand, these methods do not take into account the hierarchical structure of hand joints. As shown in Fig.~\ref{fig:pinch}, when performing gestures, the four joints of a finger have similar dynamic characteristics. The aggregation features of these four joints will be a representative description for the finger. Complex gestures generally involve the interaction among five fingers. Therefore, it is critical for gesture recognition to capture the dynamics of each finger as well as aggregate features of interactive fingers. However, it is difficult to assemble features of interactive fingers dynamically using convolution operations with fixed joints arrangements. To this end, self-attention mechanism is introduced for feature aggregation instead of convolution in this paper. With dynamic attention weights, self-attention module is able to capture the features of interactive fingers flexibly. Compared with convolution, the self-attention mechanism can directly capture global features without stacking a deep network, which is allowed to design a lightweight network for gesture recognition.

Specifically, we propose an efficient hierarchical self-attention network (HAN) for skeleton-based gesture recognition, which is based on pure self-attention without any CNN, RNN or GCN operators. The framework of HAN is shown in Fig.~\ref{fig:framework}. Four self-attention modules are designed to capture spatial and temporal features for the skeleton sequence gradually. For the skeleton at a specific frame, the skeleton joints are divided into 6 parts (5 fingers and the palm). Our method applies the joint self-attention module (\mbox{J-Att}) to extract the features of each finger. Then the finger self-attention module (\mbox{F-Att}) is designed to aggregate features of the whole hand. In terms of temporal features, the temporal self-attention module (\mbox{T-Att}) is utilized to capture the temporal dynamics of the fingers and the entire hand. Finally, these features are fused by the fusion self-attention module (\mbox{Fusion-Att}) for gesture classification. Experiments show that our method achieves competitive results on three gesture recognition datasets with much lower computational complexity. Thus, our HAN can facilitate the human computer interaction field which starves for a real-time and lightweight method to be used in mobile devices with limited computing resources. 

The main contributions of this paper are as follows:
\begin{itemize}
	\item We propose a new pure self-attention based framework without any CNN, RNN or GCN operations, which has not been well studied in skeleton-based gesture recognition task before.
	\item Considering the hierarchical structure of the hand joints, our HAN can learn exact finger level semantic features, which promote the feature aggregation of interactive fingers for hand. It also allows the \mbox{T-Att} to capture temporal features of fingers to boost the final inference.
	\item With the dedicated hierarchical design, our HAN achieves competitive results with the state-of-the-art while has a significant reduction in computational complexity, especially achieves about one-fortieth FLOPs of the best two-stream approach~\cite{liu2020decoupled}.
\end{itemize}	

The rest of the paper is organized as follows, Section~\ref{sec2} reviews the related works of skeleton-based gesture and action recognition as well as the recent works applying Transformer to computer vision tasks. In Section~\ref{sec3}, the framework and network details of HAN are presented, the self-attention module is formulated in detail. In Section~\ref{sec4}, comparison experiments, ablation study and the experimental analyses are provided. Section~\ref{sec5} draws the conclusion.

\section{Related Works}
\label{sec2}

In this section, the recent related works of skeleton-based gesture recognition are reviewed. Since the skeleton-based action recognition task is analogous with skeleton-based gesture recognition task, the papers for skeleton-based action recognition are summarized as well. We also review the latest works which introduce pure self-attention mechanism into computer vision fields.

\subsection{Skeleton-Based Gesture Recognition}
Recently, skeleton data has attracted the attention of researchers, since skeleton data provides rich and high-level information for the hand. Many works try to solve the dynamic hand gesture recognition problem using skeleton data.

De Smedt \textit{et al.}~\cite{de2016skeleton} design some effective descriptors for hand using hand joints provided by Intel RealSense depth camera. These hand-crafted descriptors represent the geometric shape of the hand. Each descriptor is  encoded by a Fisher Vector representation obtained using a Gaussian Mixture Model. The temporal features of gestures are encoded using temporal pyramid. Finally, a linear Support Vector Machine (SVM) classifier is used for gesture classification.

Chen \textit{et al.}~\cite{chen2017motion} propose to learn motion features of the hand gesture via recurrent neural network (RNN). They devise finger motion features and global motion features to describe finger movements and global movement of hand skeleton respectively. Then a bidirectional recurrent neural network is utilized to augment the motion features as well as recognize the hand gestures.

Instead of representing a 3D pose directly by its joint positions, Weng \textit{et al.}~\cite{weng2018deformable} introduce a Deformable Pose Traversal Convolution Network for representation of 3D pose. The traversal convolution optimizes the convolution kernel for each joint with various weights, which enhances the robustness to the noisy joints. The pose feature is fed into the Long Short Term Memory (LSTM) networks for gesture recognition.

Hou \textit{et al.}~\cite{hou2018spatial} present a Spatial-Temporal Attention Residual Temporal Convolutional Network to learn different levels of attention for spatial-temporal feature. With the attention mechanism, the network has the ability to concentrate on the informative time frames and features. Similarly, Devineau \textit{et al.}~\cite{devineau2018deep} use Convolutional Neural Network (CNN) for skeleton based gesture recognition as well. They propose a parallel convolutional neural network to process sequences of hand skeletal joints parallelly. The feature vectors of all channels are concatenated into a feature vector and followed by a Multi Layer Perceptron (MLP) for gesture classification. Nunez \textit{et al.}~\cite{nunez2018convolutional} also introduce a  deep learning-based
approach combining CNN and LSTM.

Chen \textit{et al.}~\cite{chen2019construct} devise a Dynamic Graph-Based Spatial-Temporal Attention method for hand gesture recognition. They construct a fully-connected graph for all hand joints. Then, the node features and edges are learned with self-attention.

Nguyen \textit{et al.}~\cite{nguyen2019neural} propose a neural network based on SPD manifold learning for skeleton-based hand gesture recognition. They first use a convolutional layer to learn discriminative features of joints. Spatial and temporal features are then extracted via spatial and temporal Gaussian aggregation. Finally, an SPD matrix is learned for gesture representation and recognition.

To decouple the gesture into hand posture variations and hand movements, Liu \textit{et al.}~\cite{liu2020decoupled} present a two-stream network for skeleton-based hand gesture recognition. The skeleton sequence is modeled
into a 3D hand posture evolution volume to describe hand posture variations. Joint movements are arranged as a 2D hand movement map. Then the hand posture variation and hand movement features are aggregated by a 3D CNN stream and a 2D CNN stream respectively.

\subsection{Skeleton-Based Action Recognition}

Skeleton-based human action recognition task is similar to skeleton-based hand gesture recognition. Some of methods for skeleton-based action recognition can be generalized to skeleton-based hand gesture recognition. In this section, we review some related works for skeleton-based action recognition. Previous works for skeleton-based action recognition can be divided into two categories: the hand-crafted based and the deep learning based methods. 

Hand-crafted based approaches aim to design different descriptors to represent human action. For instance, relative positions between pairs of joints~\cite{fan2016action}, joint angles~\cite{ohn2013joint} and motion trajectories~\cite{devanne20143}  are selected to describe actions. Evangelidis \textit{et al.}~\cite{evangelidis2014skeletal} propose to use skeletal quad for action recognition. While 3D geometric relationships of body parts in a Lie group are used in paper~\cite{vemulapalli2014human}. After hand-crafted features have been established, different types of traditional classifiers such as multi-class Support Vector Machine (SVM), Hidden Markov models (HMM), and Conditional Random Fields (CRF) are used to classify human actions. Due to the hand-crafted features, these methods suffer from weak generalization capability.

With the development of deep learning, many works apply deep networks to skeleton-based action recognition in an end to end manner. The most popular models are RNN and CNN. RNN-based methods usually represent the human action as a vector sequence, where the elements of the vector are human body joint coordinates~\cite{du2015hierarchical,song2017end,zhang2017view,li2018independently,si2018skeleton,wang2017modeling}. RNN is utilized to extract spatial-temporal features from the vector sequence. Du \textit{et al.}~\cite{du2015hierarchical} divide the human skeleton into different parts according to human physical structure, then skeleton temporal sequences are fed into hierarchical RNN to gradually extract high-level representations for skeleton-based action recognition. Song \textit{et al.}~\cite{song2017end} build a spatio-temporal attention model based on LSTM, which can adaptively focus on discriminative joints of body skeleton and learn various attention weights for different frames. Zhang \textit{et al.}~\cite{zhang2017view} devise a view adaptive RNN
for human action recognition from skeleton data, which enables the network to 
transform the skeletons of various views to much more consistent viewpoints. Instead of applying vanilla RNN, Li \textit{et al.}~\cite{li2018independently} propose an independently RNN. This new type of RNN can learn long-term dependencies while avoiding the gradient exploding and vanishing problems. Si \textit{et al.}~\cite{si2018skeleton} design a spatial reasoning network to capture spatial structural information using residual graph neural network, and an LSTM based temporal stack learning network to extract temporal dynamics of skeleton sequences.

Arranging the raw skeleton sequence into a pseudo image, many works use CNN to extract spatial and temporal features for human actions~\cite{kim2017interpretable,cao2018skeleton,banerjee2020fuzzy,liu2017enhanced}. Kim \textit{et al.}~\cite{kim2017interpretable} present the temporal convolutional neural networks for 3D human action recognition using one-dimensional residual CNN. By modeling skeleton sequence into an image, Cao \textit{et al.}~\cite{cao2018skeleton} solve the skeleton-based action recognition problem as an image classification task using gated convolutional networks which guarantee the information propagation across multiple residual blocks. Banerjee \textit{et al.}~\cite{banerjee2020fuzzy} present four feature representations to describe skeleton sequence, each of which is encoded into a single channel gray-scale image. High-level features are learned by four CNNs and the final decision score is adaptively generated via fuzzy combination of the outputs of the CNNs. To cope with viewpoint variations, Liu \textit{et al.}~\cite{liu2017enhanced} develop a view invariant transform algorithm for skeleton sequence, and visualize the skeletons as a series of color images. With visual and motion enhancement, these images are fed into multi-stream CNNs for action recognition.

Although RNN and CNN based methods achieve good performance, the approach to model skeleton sequence into coordinate vector or pseudo image is not an optimal solution, since the skeleton data is naturally embedded in a graph structure. With the advancement of deep learning on graph, many researchers concentrate on applying graph neural network to skeleton-based action recognition. Yan \textit{et al.}~\cite{yan2018spatial} embed skeleton sequence into a spatio-temporal graph, where spatial edge is in line with the natural structure of human body and temporal edge connects the two same joints between adjacent frames. Then a spatial temporal graph convolutional network (ST-GCN) is proposed to extract spatial and temporal features of skeleton sequence for action classification. Based on this work, a variety of methods have been proposed to improve recognition performance using GCN~\cite{shi2019two,li2019actional,kong2021symmetrical,si2019attention}. Shi \textit{et al.}~\cite{shi2019two} devise a novel two-stream adaptive graph convolutional network for skeleton-based action recognition, where the graph topology can be learned individually. Except for joint coordinates, the length and direction of bone are used to construct another graph to exploit the second-order information for skeleton sequence. Si \textit{et al.}~\cite{si2019attention} introduce a attention enhanced graph convolutional LSTM network to capture discriminative spatial and temporal features. By employing the attention mechanism, the network is able to enhance the information of key joints in each layer.

\subsection{Self-Attention Mechanism}

Self-attention mechanism is a basic block of Transformer~\cite{vaswani2017attention}, which is first proposed for machine translation. Since then, Transformer-based methods have attracted the attention of researchers in natural language processing (NLP), and a large number of Transformer-based methods have emerged and achieved excellent performance in NLP tasks. Inspired by the success of Transformer in NLP field,  multiple works introduce Transformer into computer vision tasks. Vision Transformer~\cite{dosovitskiy2020image} applies a pure Transformer to sequences of image patches for image classification and achieves excellent performance compared to state-of-the-art convolutional networks. Viewing object detection as a direct set prediction problem, Carion \textit{et al.} propose a Transformer-based model for object detection~\cite{carion2020end}. Zheng \textit{et al.} design a powerful segmentation model, termed SEgmentation TRansformer~\cite{zheng2020rethinking}, treating semantic segmentation as a sequence-to-sequence prediction task.

\section{Proposed Approach}
\label{sec3}

\begin{figure}[!t]
	\centerline{\includegraphics[width=0.8\columnwidth]{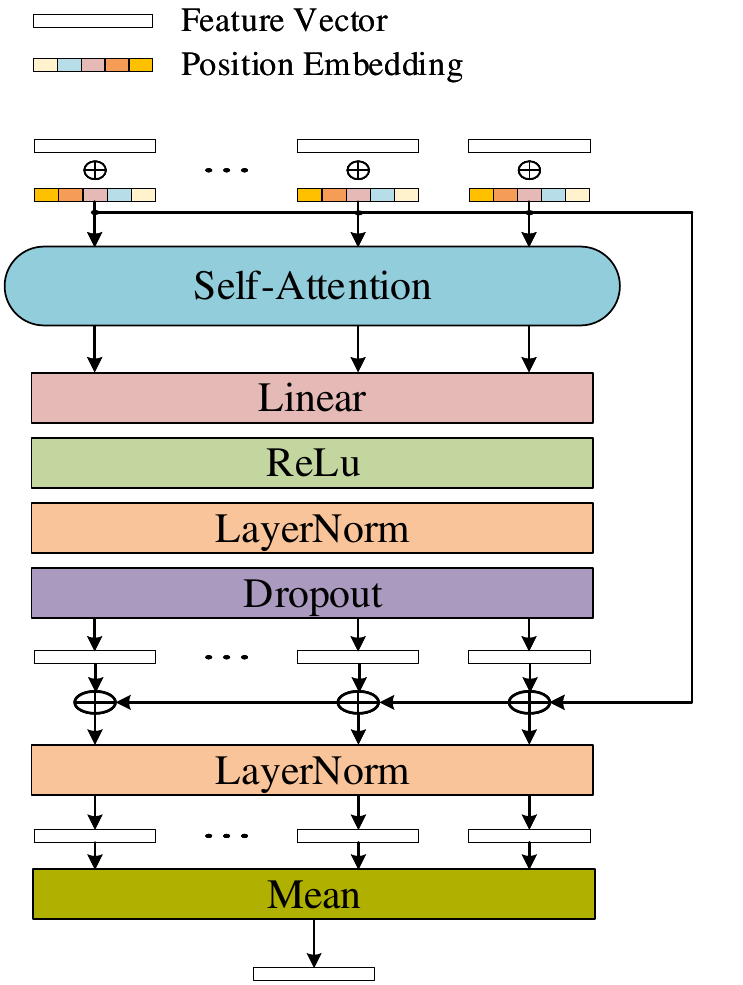}}
	\caption{Network details for attention module.}
	\label{fig:attention}
\end{figure}

The framework of HAN is shown in Fig.~\ref{fig:framework}. There are four attention modules in HAN, i.e., \mbox{J-Att}, \mbox{F-Att}, \mbox{T-Att}, and \mbox{Fusion-Att}. These four attention modules have a similar structure as shown in Fig.~\ref{fig:attention}. The attention module will be formulated detailedly taking \mbox{J-Att} as an example.

\subsection{Notations}
In the skeleton-based hand gesture recognition task, dynamic hand gesture is represented by a hand skeleton sequence. For a specific hand skeleton sequence $ G $, it can be formulated as $ G=\{S_t|t=1,2,\dots,T\} $, where $ S_t $ is the hand skeleton at $ t $-th frame, $ T $ is the length of this hand skeleton sequence. Hand skeleton is a combination of 3D coordinates of hand joints, {\em i.e.}, $ S_t=\{\textbf{p}_i^t|\textbf{p}_i^t=(x_i^t,y_i^t,z_i^t),i=1,2,\dots,J\} $, where $ \textbf{p}_i^t $ is the 3D coordinate of the $ i $-th hand joint at $ t $-th frame, and $ J $ is the number of hand joints. Given a skeleton sequence $ G $ with gesture label $ c $, the gesture recognition task aims to learn a function $ \mathcal{F}_{\varTheta}(\cdot) $ which can identify the gesture category label.

\subsection{Joint Self-Attention Module}
In hand skeleton, there are 4 joints for each finger representing the tip, the 2 articulations and the base~\cite{de2017shrec}. These four joints are similar in dynamics when performing gestures. The aggregation feature of these four joints will be an ideal representation for the finger dynamics. Hence, we design a joint self-attention (\mbox{J-Att}) module with self-attention mechanism to aggregate information of four joints for each finger. Specifically, 22 joints in each skeleton are divided into 6 parts as shown in Fig.~\ref{fig:framework}, {\em i.e.}, five fingers and the palm (the palm and the wrist). For simplicity, let's use the thumb as an example to formulate the joint self-attention module. Firstly, we apply a linear transformation to the raw joint coordinates according to Eq.~\ref{eq:linear-map},

\begin{equation}
\label{eq:linear-map}
\textbf{f}_{thu,i}^{t}=\textbf{W}_{joint} \cdot \textbf{p}_{thu,i}^{t} + b_{joint},
\end{equation}

\noindent where $ \textbf{p}_{thu,i}^{t} $ is the 3D coordinate of $ i $-th thumb joint at frame $ t $, $ i=1,2,3,4 $, $ \textbf{W}_{joint} $ and $ b_{joint} $ are parameters, $ \textbf{f}_{thu,i}^{t} $ is the corresponding output feature. To incorporate spatial identity information, $ \textbf{f}_{thu,i}^{t} $ is added with spatial position embedding vector $ PE(i) $, whose values are set using sine and cosine functions of different frequencies following~\cite{vaswani2017attention},

\begin{equation}
\label{eq:position-embed}
\tilde{\textbf{f}}_{thu,i}^{t}=\textbf{f}_{thu,i}^{t} + PE(i),
\end{equation}

\noindent where $ PE(i) $ is spatial position embedding vector for the $ i $-th thumb joint, whose dimension is same as $ \textbf{f}_{thu,i}^{t}  $.

The proposed method consists of several self-attention modules for feature aggregation. These self-attention modules share the same components, we take thumb joint attention module for illustration. For thumb joints, $ \tilde{\textbf{f}}_{thu,i}^{t} $ is mapped into key, query and value vectors using three fully connected layers, which can be formulated as:

\begin{equation}
\label{eq:KQV}
\textbf{K}_i = \textbf{W}_K \cdot \tilde{\textbf{f}}_{thu,i}^{t},  \textbf{Q}_i = \textbf{W}_Q \cdot \tilde{\textbf{f}}_{thu,i}^{t}, \textbf{V}_i = \textbf{W}_V \cdot \tilde{\textbf{f}}_{thu,i}^{t},
\end{equation}

\noindent where$ \textbf{K}_i $, $ \textbf{Q}_i $, $ \textbf{V}_i $ are key, query and value for the $ i $-th thumb joint, $ \textbf{W}_K $, $ \textbf{W}_Q $, $ \textbf{W}_V $ are the corresponding weight matrices. Then the attention weight $ \lambda_{i,j} $ between joint $ i $ and joint $ j $ is calculated by:

\begin{equation}
\label{eq:attention-weight}
u_{i,j} = \frac{\textbf{Q}_i \cdot \textbf{K}_j} {\sqrt{d}}, \lambda_{i,j} = \frac {\exp(u_{i,j})} {\sum_{n=1}^{N} \exp(u_{i,n})},
\end{equation}

\noindent where $ d $ is the dimension of key, query and value, $ N = 4 $ corresponds to four thumb joints. The self-attention feature for $ \tilde{\textbf{f}}_{thu,i}^{t} $ is defined as the weighted sum of all value vectors:

\begin{equation}
\label{eq:weight-sum}
\bar{\textbf{f}}_{thu,i}^{t} = \sum_{j=1}^{N} \lambda_{i,j} \cdot \textbf{V}_j.
\end{equation}

\noindent  In order to increase non-linearity, a fully connected layer following ReLu and layer normalization is applied to $ \bar{\textbf{f}}_{thu,i}^{t} $. With employing residual connection and dropout, the $ \bar{\textbf{f}}_{thu,i}^{t} $ is turned into:

\begin{equation}
\label{eq:attention-map}
\resizebox{.91\linewidth}{!}{$	
	\hat{\textbf{f}}_{thu,i}^{t} = \tilde{\textbf{f}}_{thu,i}^{t} + Drop(Norm(ReLu(\textbf{W}_{A} \cdot \bar{\textbf{f}}_{thu,i}^{t} + b_{A}))), 
	$}
\end{equation}

\noindent where $ \textbf{W}_{A} $ and $ b_{A} $ are parameters, $ Drop $ and $ Norm $ represent dropout and layer normalization. Finally, the attention features $ \hat{\textbf{f}}_{thu,i}^{t} (i=1,2,\cdots,N) $ are aggregated into a feature of thumb using global average pooling:

\begin{equation}
\label{eq:avg-pool}
\mathring{\textbf{f}}_{thu}^{t} = Mean(\hat{\textbf{f}}_{thu,1}^{t}, \hat{\textbf{f}}_{thu,2}^{t}, \dots, \hat{\textbf{f}}_{thu,N}^{t}),
\end{equation}

\noindent where $ \mathring{\textbf{f}}_{thu}^{t} $ is the feature describing thumb at frame $ t $. For simplification, the above operations translating $ \tilde{\textbf{f}}_{thu,i}^{t} $ to $ \mathring{\textbf{f}}_{thu}^{t} $  are denoted as Eq.~\ref{eq:attention}, $ \mathcal{A} $ denotes the self-attention module,

\begin{equation}
\label{eq:attention}
\mathring{\textbf{f}}_{thu}^{t} = \mathcal{A}(\tilde{\textbf{f}}_{thu,1}^{t}, \tilde{\textbf{f}}_{thu,2}^{t}, \cdots, \tilde{\textbf{f}}_{thu,N}^{t}).
\end{equation}

\subsection{Finger Self-Attention Module}
Through the \mbox{J-Att} module, we can obtain the spatial features of each finger, and further we use the finger self-attention (\mbox{F-Att}) module to fuse finger features to obtain the spatial features of the entire hand. Here we denote the features of the five fingers and palm as $ \textbf{f}_{hand,i}^{t}, i=1,2,\dots,N $, where $ N=6 $. $ \textbf{f}_{hand,i}^{t} $ is the output of \mbox{J-Att} module for the $ i $-th finger at $ t $ frame, {\em i.e.}, $ \textbf{f}_{hand,1}^{t} =  \mathring{\textbf{f}}_{thu}^{t} $. In order to distinguish five fingers and palm, the spatial position embedding vector $ PE(i) $ of the $ i $-th finger is added to $ \textbf{f}_{hand,i}^{t} $:

\begin{equation}
\label{eq:finger-position-embed}
\tilde{\textbf{f}}_{hand,i}^{t}=\textbf{f}_{hand,i}^{t} + PE(i).
\end{equation}

\noindent Then we use the \mbox{F-Att} module to integrate the features of each finger as the spatial features of the entire hand:

\begin{equation}
\label{eq:finger-attention}
\mathring{\textbf{f}}_{hand}^{t}=\mathcal{A}(\tilde{\textbf{f}}_{hand,1}^{t}, \tilde{\textbf{f}}_{hand,2}^{t}, \cdots, \tilde{\textbf{f}}_{hand,N}^{t}),
\end{equation}

\noindent where $ \mathcal{A} $ is the self-attention module described in the previous section, $ \mathring{\textbf{f}}_{hand}^{t} $ represents the spatial features of the whole hand at frame $ t $.

\subsection{Temporal Self-Attention Module}
In order to obtain the temporal dynamics of each finger and the entire hand, we use the temporal self-attention (\mbox{T-Att}) module to aggregate the spatial features of the fingers and hand at different times. Still taking the thumb as an example, $ \mathring{\textbf{f}}_{thu}^{t} $ represents the spatial feature of the thumb at $ t $ frame, where $ t=1,2,\cdots,T $. To distinguish the chronological order, we add a time position embedding vector to $ \mathring{\textbf{f}}_{thu}^{t} $:

\begin{equation}
\label{eq:time-position-embed}
\tilde{\textbf{f}}_{thu}^{t}=\mathring{\textbf{f}}_{thu}^{t} + PE(t).
\end{equation}

\noindent Similarly, the \mbox{T-Att} module is employed to integrate the temporal dynamics of the thumb:

\begin{equation}
\label{eq:time-attention}
\mathring{\textbf{f}}_{thu}=\mathcal{A}(\tilde{\textbf{f}}_{thu}^{1}, \tilde{\textbf{f}}_{thu}^{2}, \cdots, \tilde{\textbf{f}}_{thu}^{T}).
\end{equation}

\subsection{Fusion Self-Attention Module}
So far we have obtained the temporal dynamics of the fingers and the entire hand. Finally, we design a fusion self-attention (\mbox{Fusion-Att}) module to fuse these temporal dynamic features, the fusion features is utilized to infer the gesture category of the skeleton sequence. We denote the temporal dynamics of each part as $ \textbf{f}_{i}, i=1,2,\dots,N $, where $ N=7 $ corresponds to 5 fingers, palm and the entire hand. Position embedding vector is added to $ \textbf{f}_{i} $ to distinguish these features as well:

\begin{equation}
\label{eq:fusion-position-embed}
\tilde{\textbf{f}}_{i}={\textbf{f}}_{i}+ PE(i).
\end{equation}

\noindent \mbox{Fusion-Att} is then used to fuse these temporal dynamics:

\begin{equation}
\label{eq:fusion-attention}
\mathring{\textbf{f}}=\mathcal{A}(\tilde{\textbf{f}}_{1}, \tilde{\textbf{f}}_{2}, \cdots, \tilde{\textbf{f}}_{N}),
\end{equation}

\noindent where $ \mathring{\textbf{f}} $ is the spatial-temporal features of the whole skeleton sequence. As shown in Figure~\ref{fig:framework}, this feature vector is followed by a fully connected layer with a softmax activation for gesture recognition.

\section{Experiments}
\label{sec4}

In this section, three datasets and the implementation details of HAN are introduced. The performance of our approach is compared with the state-of-the-art approaches. The complexity comparison and ablation study are presented. The attention visualization and the hyper-parameters are discussed.

\begin{table}
	\centering
	\caption{Attributes of three gesture datasets.}
	\begin{tabular}{c||ccc} 
		\toprule
		Dataset & SHREC'17 Track & DHG-14/28 & FPHA  \\ 
		\hline
		Gestures & 14/28 & 14/28 & 45  \\ 
		\hline
		Subjects & 28 & 20 & 6  \\ 
		\hline
		Perform & 1-10 & 5 &  4-5 \\ 
		\hline
		Sequences & 2800 & 2800 & 1175  \\
		\hline
		Color & \xmark & \xmark & \cmark  \\
		\hline
		Depth & \cmark & \cmark & \cmark  \\
		\hline
		Skeleton & \cmark & \cmark & \cmark  \\
		\hline
		Joints & 22 & 22 & 21  \\
		\bottomrule
	\end{tabular}
	\label{tab:datasets}
\end{table}

\begin{figure}[!t]
	\centerline{\includegraphics[width=\columnwidth]{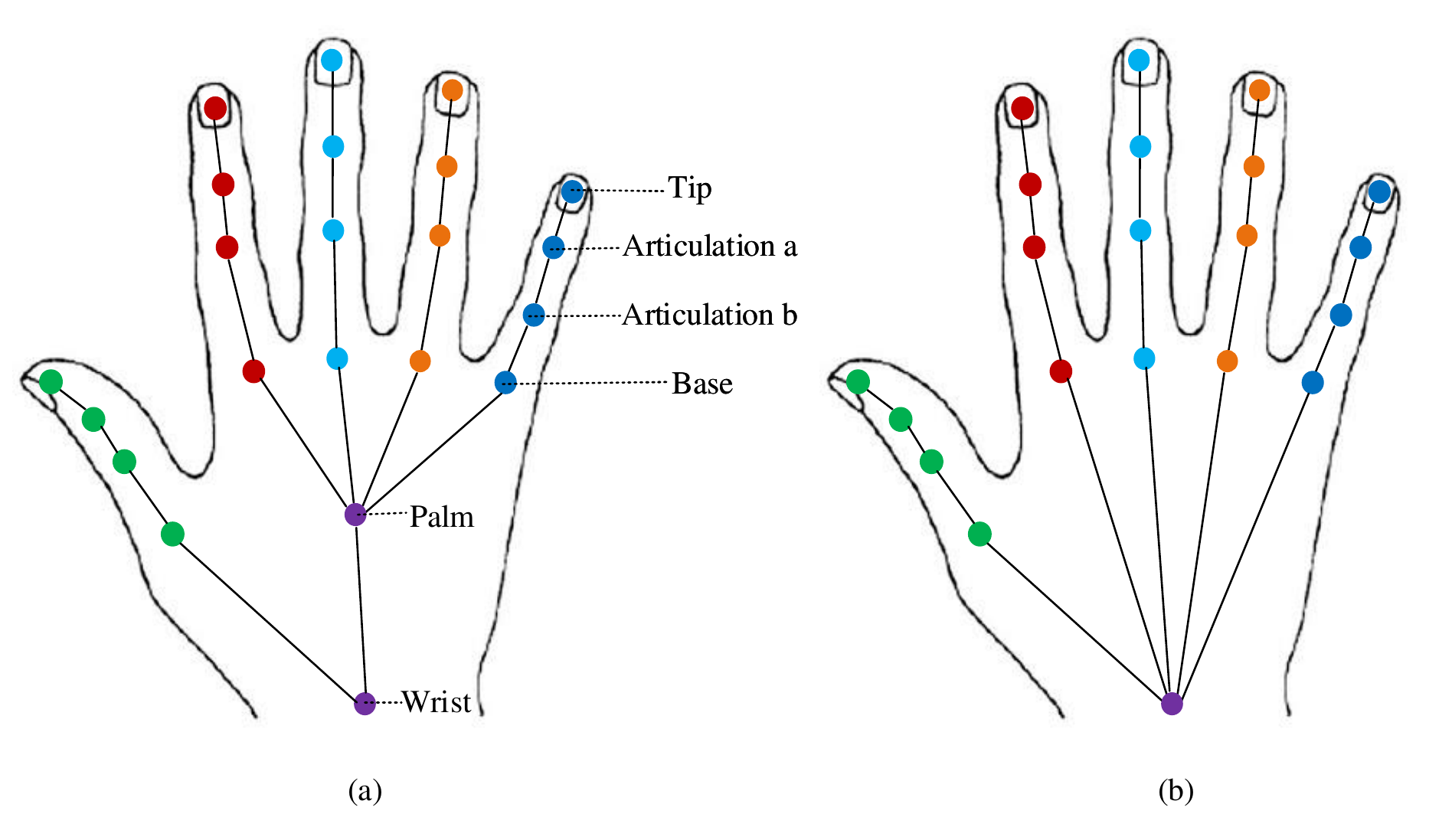}}
	\caption{The hand skeleton of different datasets. (a) Hand skeleton on SHREC'17 Track and DHG-14/28 dataset with 22 hand joints. (b) Hand skeleton on FPHA dataset with 21 hand joints excluding palm joint.}
	\label{fig:hand-skeleton}
\end{figure}

\subsection{Datasets}
The performance of our method is evaluated on three challenging gesture datasets with skeleton data: SHREC'17 Track~\cite{de2017shrec}, DHG-14/28~\cite{de2016skeleton}, FPHA~\cite{garcia2018first}.  In this subsection, we introduce these three datasets in detail. The attributes of these datasets are listed in Table~\ref{tab:datasets}. The hand skeleton joints of these datasets are illustrated in Fig~\ref{fig:hand-skeleton}.

\subsubsection{SHREC'17 Track} 
The SHREC'17 Track dataset~\cite{de2017shrec} is a challenging hand gesture dataset which provides both depth images and skeleton data. Collected by Intel RealSense short range depth camera, SHREC'17 Track dataset contains 14 gestures performed by 28 individuals that are all right-handed. In order to meet the accuracy requirements of human-computer interaction in AR and VR applications, hand gesture recognition algorithms must take into account the difference between hand movement and hand shape changes. Hence, this dataset divides 14 gestures into two categories, fine (5 out of 14) and coarse (9 out of 14). A fine gesture is one that considers hand shape changes, and a coarse gesture is a gesture that focuses on hand movement. Each gesture is performed between 1 and 10 times in two ways, using one finger and the whole hand. The sequences are captured 30 frames per second, the depth image resolution is 640$ \times $480 and each sequence ranges from 20 to 50 frames. Each frame contains a depth image and coordinates of 22 joints under 2D and 3D spaces. In the end, 2800 sequences in all are collected, which are divided into 1960 sequences (70\% of the dataset) for training and 840 sequences (30\% of the dataset) for testing.

\subsubsection{DHG-14/28} 
Also resulting in 2800 sequences of 22 hand joints, the DHG-14/28 dataset~\cite{de2016skeleton} uses the same data collection method as the SHREC'17 Track dataset. While the DHG-14/28 dataset is performed by 20 individuals rather than 28. Each gesture is performed exactly 5 times instead of a random number between 1 to 10. The main difference between the two datasets is the experimental protocol. Comparing with the SHREC'17 Track dataset, the DHG-14/28 dataset does not have an explicit partition for training set and testing set. Leave-one-subject-out experimental protocol is used for the DHG-14/28 dataset~\cite{de2016skeleton, de2019heterogeneous, nguyen2019neural, chen2019construct}. During training, the data of 19 subjects is regarded as training set, the data of the leaving one subject is used for testing. Experiments are conducted 20 times with different subjects for testing, the average accuracy is considered as final recognition accuracy.

\subsubsection{FPHA} 
Unlike SHREC'17 Track dataset and DHG-14/28 dataset, FPHA dataset~\cite{garcia2018first} is collected from a first-person perspective. The FPHA dataset is performed in 3 scenarios which are kitchen, office and social, by 6 individuals, providing 1175 action videos that can be categorized into 45 different daily hand actions involving 26 objects. Using an Intel RealSense SR300 RGB-D camera on the subject's shoulder, together with six magnetic sensors attached to the user’s hand (five on the fingertips and one on the wrist), skeleton sequences are captured at 30 fps as well as the color and depth stream with resolutions of 1920$ \times $1080 and 640$ \times $480 respectively. Each sensor has 6 degrees of freedom of position and orientation. Through these sensors, a 21-joint hand model can be defined. Different action sequences are quite different in style, speed, scale and viewpoint, whether it is indifferent actions of the same subject or between different subjects. The dataset also provides 6-dimensional poses, 3D positions and angles, and mesh models for 4 objects involving 10 different action categories. In all, 105,459 RGB-D frames are annotated with accurate hand pose and action category. Similar to SHREC'17 Track dataset and DHG-14/28 dataset, 3D coordinates of 21 hand joints (excluding the palm joint) are provided in the FPHA dataset. As for training and testing, 600 out of 1175 action sequences are divided for training and 575 for testing.

\subsection{Implementation Details}
For all the self-attention modules, the dimension of the input feature vector and output feature vector are both set to 128. Thus all position embedding vectors are with 128 dimensions. To learn more attention information from different representation subspaces at different positions, multi-head attention~\cite{vaswani2017attention} is applied. We use 8 heads attention in our HAN. The dimensions of key, query and value vectors for each head are fixed to 32. The \mbox{J-Att} modules share weights for 6 hand parts. Similarly, the \mbox{T-Att} modules share weights for 7 inputs.

Our HAN  is implemented on a computer with one NVIDIA TITAN Xp GPU using PyTorch library. The skeleton sequence is uniformly sampled to 8 frames as an input following previous works~\cite{hou2018spatial,chen2019construct}. For a fair comparison, the same data augmentation is performed according to the paper~\cite{de2017shrec}, including scaling, shifting, time interpolation, and adding noise. Adam is chosen as the optimization strategy and cross-entropy is selected as the loss function. The batch size is set to 32 for training and the dropout rate is fixed to 0.1. The learning rate starts from 0.001 with warm-up epochs and decays by a factor of 10 once learning stagnates. The training process is stopped when the learning rate decays for the fourth time.

\begin{table}[!t]
	\centering
	\caption{Performance comparison with the state-of-the-art approaches on SHREC'17 Track dataset with 1960 sequences for training and 840 sequences for testing. 14G and 28G represent 14 and 28 gesture settings. The first row contains the two-stream methods, the second row involves the single-stream methods, and the third row is our method.}
	\label{tab:SHREC-comparsion}
	\begin{tabular}{l|cc}
		\toprule
		\multirow{2}*{\textbf{Method}} & \multicolumn{2}{c}{\textbf{Accuracy (\%)}} \\
		\cline{2-3}
		&\textbf{14G} & \textbf{28G} \\		
		\hline
		Two-stream 3D CNN~\cite{tu2018skeleton} & 83.45 & 77.43\\
		SEM-MEM+WAL~\cite{liu2018learning} & 90.83 & 85.95 \\
		Liu {\em et al.}~\cite{liu2020decoupled} & 94.88 & 92.26 \\
		\hline
		HON4D~\cite{oreifej2013hon4d} & 78.50 & 74.00 \\
		Devanne {\em et al.}~\cite{devanne20143} & 79.40 & 62.00 \\
		Ohn-Bar {\em et al.}~\cite{ohn2013joint} & 83.90 & 76.50 \\
		SoCJ+Direction+Rotation~\cite{de2017dynamic} & 86.90 & 84.20 \\
		SoCJ+HoHD+HoWR~\cite{de2016skeleton} & 88.20 & 81.90 \\		
		Caputo {\em et al.}~\cite{caputo2018comparing} & 89.
		50 & - \\
		Boulahia {\em et al.}~\cite{boulahia2017dynamic}  & 90.50 & 80.50 \\
		Res-TCN~\cite{hou2018spatial} & 91.10 & 87.30 \\
		STA-Res-TCN~\cite{hou2018spatial} & 93.60 & 90.70 \\
		ST-GCN~\cite{yan2018spatial} & 92.70 & 87.70 \\
		ST-TS-HGR-NET~\cite{nguyen2019neural} & 94.29 & 89.40 \\
		DG-STA~\cite{chen2019construct}  & 94.40 & 90.70  \\
		\hline
		\textbf{HAN} & \textbf{95.00} & 91.07 \\
		\textbf{HAN-2S} & \textbf{95.00} & \textbf{92.86} \\	
		\bottomrule
	\end{tabular}
\end{table}

\subsection{Experiments on SHREC'17 Track Dataset}

The SHREC'17 Track dataset is a challenging skeleton-based hand gesture dataset. It contains 2800 skeleton sequences with 14 gestures performed by 28 individuals in two ways: using one finger and the whole hand. The dataset is divided into 1960 sequences for training and 840 sequences for testing. We compare our approach with various typical works for gesture recognition on SHREC'17 Track dataset. These methods can be divided into six categories: 

\begin{itemize}
	\item \textbf{Hand-crafted methods}. HON4D\cite{oreifej2013hon4d}: It represents the depth sequence of action using a histogram describing the distribution of the surface normal orientation in the 4D space. Devanne~\textit{et al.}~\cite{devanne20143}: They formulate the problem of action recognition as computing the similarity between the shape of 3D human joint trajectories in a Riemannian manifold. Ohn-Bar~\textit{et al.}~\cite{ohn2013joint}: They propose to characterize actions using joint angles and modified histogram of oriented gradients. SoCJ+Direction+Rotation~\cite{de2017dynamic}: It describes hand gestures via hand shape and motion descriptors from 3D hand skeleton data. SoCJ+HoHD+HoWR~\cite{de2016skeleton}: It represents the geometric shape of the hand using hand-crafted descriptor which is encoded by a Fisher Vector representation obtained using a Gaussian Mixture Model. Caputo~\textit{et al.}~\cite{caputo2018comparing}: They recognize hand gestures by gesture traces processing and comparing. Boulahia~\textit{et al.}~\cite{boulahia2017dynamic}: They present a dynamic hand gesture recognition method based on 3D pattern assembled trajectories.
	
	\item \textbf{CNN-based methods}. SEM-MEM+WAL~\cite{liu2018learning}: It represents actions using shape and motion evolution maps. STA-Res-TCN~\cite{hou2018spatial}: It learns different levels of attention for spatial-temporal features using a residual temporal convolutional network. Res-TCN~\cite{hou2018spatial}: An ablation version of STA-Res-TCN~\cite{hou2018spatial} without spatial-temporal attention.
	
	\item \textbf{3D-CNN-based methods}. Two-stream 3D CNN~\cite{tu2018skeleton}: It proposes a spatial and temporal volume modeling approach for skeleton sequence. Liu {\em et al.}~\cite{liu2020decoupled}: They decouple the gesture into hand posture variations and hand movements and embed them into hand posture variation volume and hand movement map respectively.
	
	\item \textbf{Graph-based method}. ST-GCN~\cite{yan2018spatial}: It embeds the skeleton sequence into a spatial-temporal graph.
	
	\item \textbf{Attention-based method}. DG-STA~\cite{chen2019construct}: It constructs dynamic graphs for hand gesture recognition via spatial-temporal attention.
	
	\item \textbf{Manifold-learning-based method}. ST-TS-HGR-NET\cite{nguyen2019neural}: It proposes a neural network based on SPD manifold learning for skeleton-based hand gesture recognition.	
\end{itemize}

\begin{table}[!t]
	\centering
	\caption{Performance comparison with the state-of-the-art methods on DHG-14/28 dataset using the leave-one-subject-out experimental protocol. 14G and 28G represent 14 and 28 gesture settings. The first row contains the two-stream methods, the second row involves the single-stream methods, and the third row is our method}
	\label{tab:DHG-comparsion}
	\begin{tabular}{l|cc}
		\toprule
		\multirow{2}*{\textbf{Method}} & \multicolumn{2}{c}{\textbf{Accuracy (\%)}} \\
		\cline{2-3}
		&\textbf{14G} & \textbf{28G} \\
		\hline
		Liu {\em et al.}~\cite{liu2020decoupled} & 92.54 & 88.86 \\
		\hline
		SoCJ+HoHD+HoWR~\cite{de2016skeleton} & 83.10 & 80.00 \\
		Chen {\em et al.}~\cite{chen2017motion} & 84.70 & 80.30 \\
		CNN+LSTM~\cite{nunez2018convolutional} & 85.60  & 81.10 \\
		Weng {\em et al.}~\cite{weng2018deformable} & 85.80 & 80.40 \\
		Res-TCN~\cite{hou2018spatial} & 86.90 & 83.60 \\
		STA-Res-TCN~\cite{hou2018spatial} & 89.20 & 85.00 \\
		ST-GCN~\cite{yan2018spatial} & 91.20 & 87.10 \\
		ST-TS-HGR-NET~\cite{nguyen2019neural} & 87.30 & 83.40 \\
		DG-STA~\cite{chen2019construct}  & 91.90 & 88.00 \\
		\hline
		\textbf{HAN} & 92.62 & 88.83 \\	
		\textbf{HAN-2S} & \textbf{92.71} & \textbf{89.15} \\
		\bottomrule
	\end{tabular}
\end{table}

Table~\ref{tab:SHREC-comparsion} shows the comparison of different methods on SHREC'17 Track dataset. The results of other methods are collected from papers~\cite{liu2020decoupled, chen2019construct, nguyen2019neural}. The first row of the table contains the two-stream methods, and the second row contains the single-stream methods. It shows that our approach achieves a recognition accuracy of 95.00\% for 14 gestures setting and 91.07\% for 28 gestures setting. Our method achieves the best performance within the single-stream methods, and the results are also competitive compared with the best two-stream work. For a fair comparison, we evaluate the two-stream version of our method (\textbf{HAN-2S}) on the dataset following the same settings with the best two-stream method~\cite{liu2020decoupled}. Results indicate that HAN-2S performs better than the state-of-the-art two-stream method~\cite{liu2020decoupled}. It achieves a recognition accuracy of 95.00\% for 14 gestures setting and 92.86\% for 28 gestures setting. Comparing with HAN, HAN-2S has a notable improvement on performance with 28 gestures setting. We argue that the fingertip relative position feature~\cite{liu2020decoupled} is crucial for distinguishing gestures performed using one finger or the whole hand.

\subsection{Experiments on DHG-14/28 Dataset}
On DHG-14/28 dataset, experiments are conducted following leave-one-subject-out experimental protocol~\cite{de2016skeleton}. Although the DHG-14/28 dataset shares the same hand gestures with the SHREC'17 Track dataset, it is more challenging for DHG-14/28 dataset. Under leave-one-subject-out experimental protocol, the same kind of gestures performed by different actors may vary a lot.

The results comparison on DHG-14/28 dataset is listed in Table~\ref{tab:DHG-comparsion}. HAN achieves a recognition accuracy of 92.62\% with 14 gestures setting and 88.83\% with 28 gestures setting, which has a minor performance improvement compared with the best of previous approaches. With a two-stream configuration, HAN-2S obtains more performance improvements with 92.71\% under 14 gestures setting and 89.15\% under 28 gestures setting. The performance of our approach on DHG-14/28 dataset is consistent with the SHREC'17 Track dataset, which further demonstrates the effectiveness of our method.

\begin{table}[!t]
	\centering
	\caption{Recognition accuracy comparison of our method with state-of-the-art methods on FPHA dataset. The first row contains the two-stream methods, the second row involves the single-stream methods, and the third row is our method}
	\label{tab:FPHA comparsion}
	\begin{tabular}{l|c}
		\toprule
		\textbf{Method} & \textbf{Acc. (\%)} \\
		\hline
		Liu {\em et al.}~\cite{liu2020decoupled} &  90.96 \\
		\hline
		Moving Pose~\cite{zanfir2013moving} &  56.34 \\
		Lie Group~\cite{vemulapalli2014human} &  82.69 \\
		HBRNN~\cite{du2015hierarchical} &  77.40 \\
		Gram Matrix~\cite{zhang2016efficient} & 85.39 \\
		TF~\cite{garcia2017transition} & 80.69 \\
		Huang et al.~\cite{huang2017riemannian} & 84.35 \\
		Huang et al.~\cite{huang2018building} & 77.57 \\
		ST-TS-HGR-NET~\cite{nguyen2019neural} & \textbf{93.22} \\
		\hline
		\textbf{HAN} & 85.74 \\	
		\textbf{HAN-2S} & 89.04 \\	
		\bottomrule
	\end{tabular}
\end{table}

\subsection{Experiments on FPHA Dataset}

To verify the scalability of our method, contrast experiments are performed on FPHA dataset for hand action recognition. The FPHA dataset is more challenging than SHREC'17 Track dataset and DHG-14/28 dataset. It contains 45 hand actions, which is much more than 28 gestures for SHREC'17 Track dataset and DHG-14/28 dataset. Besides, the FPHA dataset only provides 600 out of 1175 action sequences for training, while 575 action sequences for testing. It is a really small dataset for deep learning method.

Results in Table~\ref{tab:FPHA comparsion} show our method achieves competitive performance. Compared with ST-TS-HGR-NET~\cite{nguyen2019neural} which is based on manifold learning with SVM classifier, our method is suboptimal. We argue that Transformer-based method may be suitable for a large dataset, or pre-training on a large dataset and fine-tune on a small dataset~\cite{dosovitskiy2020image}, while manifold learning based method with SVM classifier may obtain better performance on a small dataset due to its superior generalization ability. Actually, ST-TS-HGR-NET~\cite{nguyen2019neural} only attains 94.29\% with 14 gestures setting and 89.40\% with 28 gestures setting on SHREC'17 Track dataset, which has a major performance gap with HAN. Especially on DHG-14/28 dataset, it only obtains 87.30\% with 14 gestures setting and 83.40\% with 28 gestures setting, where our method outperforms ST-TS-HGR-NET by 5.41 and 5.75 percent points for 14 gestures and 28 gestures protocol. The inferior recognition results of ST-TS-HGR-NET on SHREC'17 Track and DHG-14/28 datasets support our analysis above.

\subsection{Complexity Comparison}
The main application area of gesture recognition is human-computer interaction in VR and AR devices with limited computing resources. Thus, a fast and lightweight method with good performance on recognition is vital for skeleton-based gesture recognition. Recent approaches applying deep CNN, RNN and GCN have shown saturated recognition accuracy without significant improvement, while the computing efficiency still remains a series issue. Our HAN is a new pure self-attention based framework for skeleton-based gesture recognition without any CNN, RNN or GCN operations. The self-attention mechanism help to aggregate global features of inputs efficiently without stacking deep networks. With the dedicated hierarchical design, our HAN achieves competitive results with SOTA while maintains much lower computational complexity of the whole network.

To investigate the computation complexity of different methods, Table~\ref{tab:complexity-comparsion} lists the computation complexity details of some typical approaches and the proposed method. The parameters and FLOPs of our HAN model with default settings are 0.53M and 0.04G respectively. Compared with other methods, HAN has fewer parameters. Especially in terms of computation complexity, the FLOPs of our method are only about one fortieth of the best two-stream approach~\cite{liu2020decoupled}. Therefore, HAN is a lightweight network which has the potential for being deployed on a terminal device with limited computing resources.

\begin{table}[!t]
	\centering
	\caption{Details of computational complexity of different skeleton-based approaches.}
	\label{tab:complexity-comparsion}
	\begin{tabular}{l|cc}
		\toprule
		{\textbf{Method}} & {\textbf{Params}} & {\textbf{FLOPs}} \\	
		\hline
		Two-stream 3D CNN~\cite{tu2018skeleton} & 0.91M & 1.53G \\
		SEM-MEM+WAL~\cite{liu2018learning} & 2.36M & 1.35G \\
		ST-GCN~\cite{yan2018spatial} & 1.73M & 1.68G \\
		Liu {\em et al.}~\cite{liu2020decoupled} & 2.05M & 1.46G \\
		\hline
		\textbf{HAN} & \textbf{0.53M} & \textbf{0.04G} \\
		\textbf{HAN-2S} & \textbf{0.94M} & \textbf{0.05G} \\	
		\bottomrule
	\end{tabular}
\end{table}

\begin{figure}[!t]
	\centerline{\includegraphics[width=0.8\columnwidth]{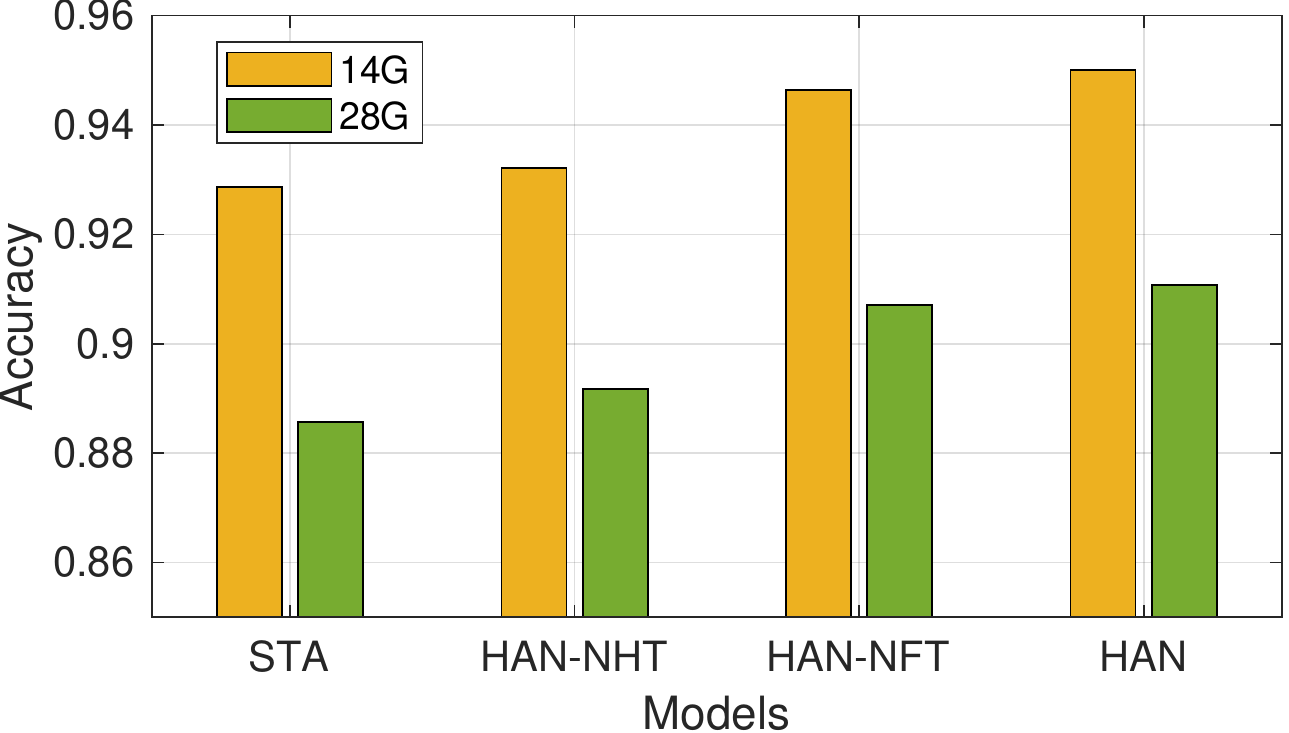}}
	\caption{Results with different model components on SHREC'17 Track dataset with 14 and 28 gestures settings.}
	\label{fig:ablation}
\end{figure}

\begin{table}[!t]
	\caption{Performance of the fine and coarse category gestures on SHREC'17 Track dataset.}
	\label{tab:fine_coarse}
	\centering
	\begin{tabular}{c|cc|cc}
		\toprule
		\multirow{2}{*}{\textbf{Method}} & \multicolumn{2}{c|}{14 Gestures} & \multicolumn{2}{c}{28 Gestures}  \\ 
		\cline{2-5}
		& Fine & Coarse	& Fine  & Coarse	\\ 
		\hline
		STA & 90.14 & 94.20	&  83.01 & 91.31	\\
		HAN-NHT & 90.59 & 94.50	& 83.56  & 91.93	\\
		HAN-NFT & 92.40 & 95.74	&  85.52 & 93.26	\\
		\textbf{HAN} & \textbf{92.78} & \textbf{96.09} & \textbf{85.92} & \textbf{93.61}\\
		\bottomrule
	\end{tabular}
\end{table}

\subsection{Ablation Study}

\subsubsection{Contribution of Different Components}
To evaluate model components, four variants are verified on SHREC'17 Track dataset with 14 and 28 gestures settings. (1) \textbf{STA}: Baseline of HAN, only spatial attention and temporal attention modules are used to capture hand spatial features and temporal dynamics without regard to hierarchical relationship of hand joints. (2) \textbf{HAN-NHT}: The \mbox{F-Att} and \mbox{T-Att} modules for the entire hand are removed. (3) \textbf{HAN-NFT}: No \mbox{T-Att} module for fingers. (4) \textbf{HAN}: The whole network of our method. The experiment results of these variants are given in Fig.~\ref{fig:ablation}. It is observed that the \mbox{T-Att} module for fingers and the entire hand can both improve performance, which verifies the effectiveness of hierarchical structure of HAN. Furthermore, combining the temporal dynamics of fingers and entire hand will achieve better results.

\begin{figure}[!t]
	\centerline{\includegraphics[width=\columnwidth]{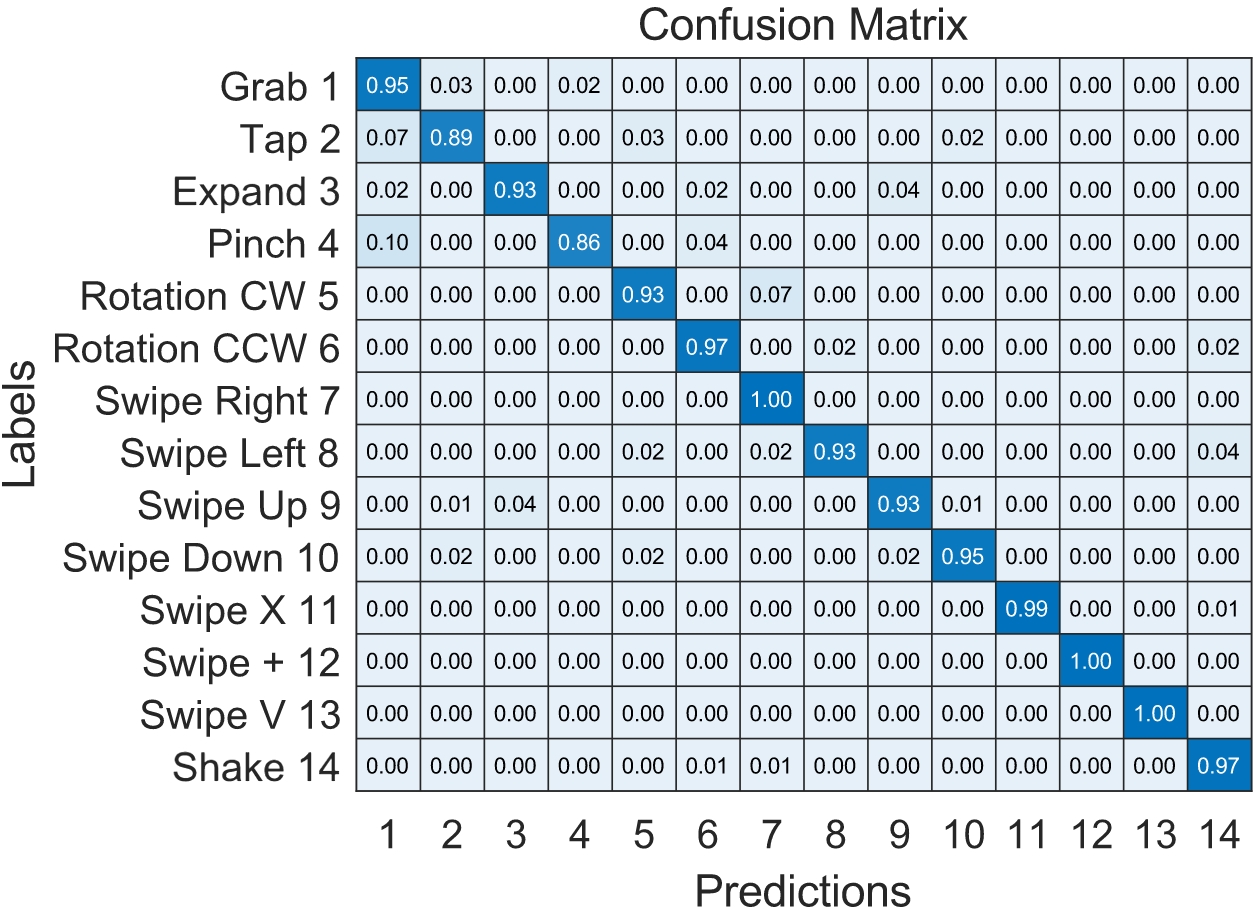}}
	\caption{The confusion matrix for the predictions of HAN on SHREC'17 Track dataset with 14 gestures setting.}
	\label{fig:ConfusionMatrix}
\end{figure}

Fig.~\ref{fig:ConfusionMatrix} shows the confusion matrix for the predictions of HAN on SHREC'17 Track dataset with 14 gestures setting. The performance on most gesture categories is over 90\%. Gesture ``Pinch'' has the worst recognition accuracy with confusion between gesture ``Pinch'' and gesture ``Grab''. These two kinds of gestures are similar in dynamics since they share the same hand shape changes. The only difference between these two gestures is the movements of the hand.

The 14 gestures of SHREC'17 Track dataset can be divided into two categories, fine gestures (5 out of 14, \textit{i.e.}, ``Grab'', ``Expand'', ``Pinch'', ``Rotation Clockwise'', and ``Rotation Counter Clockwise'') and coarse gestures (9 out of 14, \textit{i.e.}, ``Tap'', ``Swipe Right'', ``Swipe Left'', ``Swipe Up'', ``Swipe Down'', ``Swipe X'', ``Swipe +'', ``Swipe V'', and ``Shake''). A fine gesture mainly contains hand shape changes, and a coarse gesture is a gesture that focuses on hand movements. To explore the performance of STA, HAN-NFT, HAN-NHT, and HAN on the two kinds of gestures, we list the results for the fine and coarse gestures in Table~\ref{tab:fine_coarse}. Comparing with the fine gestures, the performance of the four variants on coarse gestures is much better than that on fine gestures. The results under 28 gestures setting are consistent with that under 14 gestures setting. This is because the coarse gestures only involve the movements of the entire hand without complex hand shape changes, which is easier to learn. Compared with STA, HAN has a performance improvement with 2.64\% and 1.89\% on the fine and coarse gestures. More performance improvements on fine gestures indicates that the hierarchical structure of HAN helps enhance the network's ability for capturing hand shape change features.

Without hierarchical structure, STA aggregates hand features using global attention from all hand joints directly. Thus, the hand features are easily affected by the noise hand joints. However, with hierarchical self-attention design, the finger with noise joints may obtain small weight due to its indistinctive dynamics. Besides, the hierarchical structure helps to learn exact finger level semantic features, which promote the feature aggregation of interactive fingers for hand. It also allows the \mbox{T-Att} to capture temporal features of each finger to boost the final inference.

\begin{table}[!t]
	\centering
	\caption{Ablation study for position embedding on SHREC'17 Track dataset with 14 gestures setting.}
	\label{tab:ablation-PE}
	\begin{tabular}{l|cccc|c} 
		\toprule
		& \multicolumn{4}{c|}{Ablations} & \textbf{HAN}  \\ 
		\hline
		\mbox{J-Att} PE 		& \xmark & \cmark & \cmark & \cmark & \cmark  \\
		\mbox{F-Att} PE 		& \cmark & \xmark & \cmark & \cmark & \cmark  \\
		\mbox{T-Att} PE 		& \cmark & \cmark & \xmark & \cmark & \cmark  \\
		\mbox{Fusion-Att} PE	& \cmark & \cmark & \cmark & \xmark & \cmark  \\ 
		\hline
		Accuracy(\%)	&  94.05 &  93.33 &  88.57 & 93.93  & \textbf{95.00}  \\
		\bottomrule
	\end{tabular}
\end{table}

\begin{figure*}[!t]
	\centerline{\includegraphics[width=\textwidth]{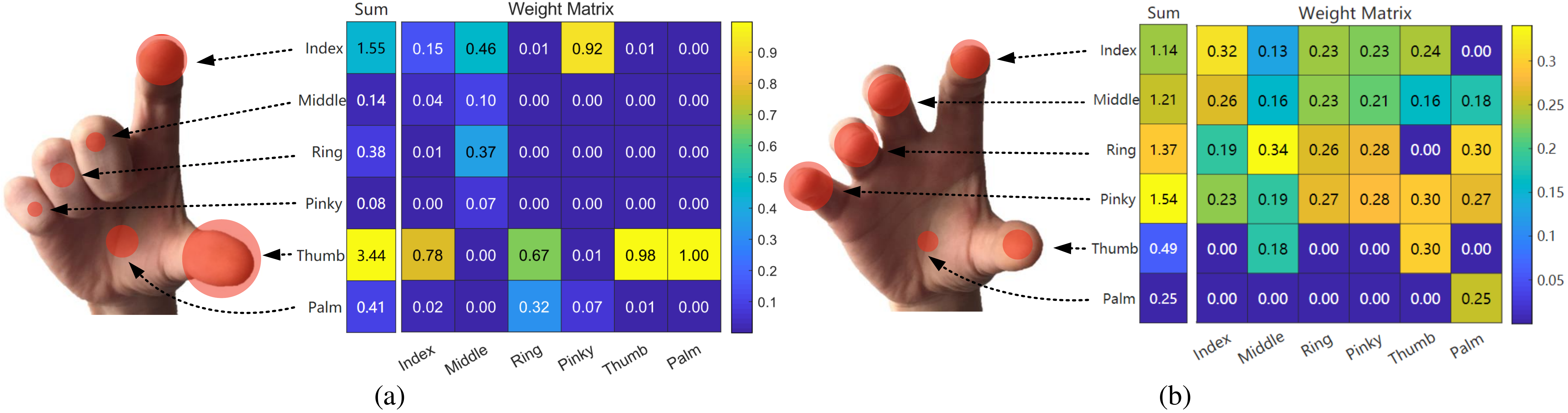}}
	\caption{Weight matrix of fingers to other fingers in \mbox{F-Att} module. The radius of the red circle on the finger represents the sum of the weights to this finger. The displayed weight matrix is an average value of multi-head attention. (a) Gesture ``Pinch'', which is performed by thumb and index fingers. (b) Gesture ``Pinch'', which is performed using the whole hand.}
	\label{fig:weightmatrix}
\end{figure*}

\begin{figure*}[!t]
	\centerline{\includegraphics[width=\textwidth]{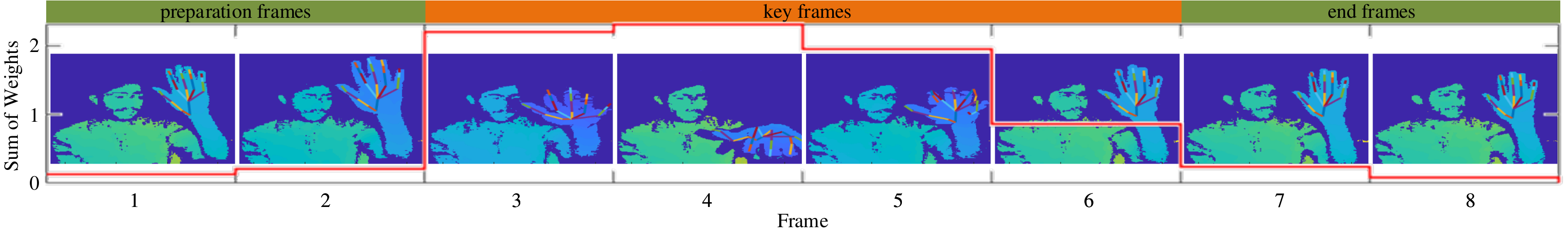}}
	\caption{Attention visualization in \mbox{T-Att} module for gesture ``Tap''. The red staircase curve describes the sum of weights to this frame.}
	\label{fig:T-Att}
\end{figure*}

\subsubsection{Influence of Position Embedding}
There are four position embeddings used in HAN. The position embedding in \mbox{J-Att} module distinguishes four joints in a finger. To tell features of which finger, the position embedding for finger features is used in \mbox{F-Att} module. The time position embedding is designed to distinguish the chronological order of skeleton sequence in \mbox{T-Att} module. In \mbox{Fusion-Att} module, the position embedding is utilized to differentiate features of different parts. In order to investigate the influence of position embedding, we conduct ablation experiments on SHREC'17 Track dataset with 14 gestures setting. The performance of models without position embedding in different modules is listed in Table~\ref{tab:ablation-PE}. The results indicate that the time position embedding in \mbox{T-Att} module is the most crucial component for gesture recognition. This is reasonable because the chronological order of the skeleton sequence is really important for gesture recognition. Without the correct chronological order, it is easy to confuse most gesture categories, such as gestures ``Swiping Left'' and ``Swiping Right''. The position embedding for joint features in \mbox{J-Att} module is not so critical for gesture recognition. We argue that the dynamics of four joints in one finger are similar to each other as shown in Fig.~\ref{fig:pinch}, thus the identities of joints are subordinate to finger feature.


\subsection{Attention Visualization}
Fig.~\ref{fig:weightmatrix} (a) shows the learned attention weight matrix of fingers to each other for gesture ``Pinch'' in \mbox{F-Att} module. When performing this gesture, the dynamics of thumb and index finger are most obvious. Fig.~\ref{fig:weightmatrix} (a) indicates that the learned attention weights are successfully concentrated on the thumb and index fingers. The gestures in SHREC'17 Track dataset are performed in two ways: using one finger and the whole hand. Fig.~\ref{fig:weightmatrix} (a) shows the attention weight for gesture ``Pinch'' which is performed using one finger. Fig.~\ref{fig:weightmatrix} (b) illustrates the attention weight matrix for gesture ``Pinch'' which is performed using the whole hand. It shows that \mbox{F-Att} learns balanced weights for the 6 parts of the whole hand. Comparing Fig.~\ref{fig:weightmatrix} (a) and Fig.~\ref{fig:weightmatrix} (b), our HAN is able to focus on the fingers with obvious dynamics.

Similarly, we illustrate the attention weights in \mbox{T-Att} module for gesture ``Tap'' in Fig.~\ref{fig:T-Att}. The sum of weights to each frame is represented as the red staircase curve with the corresponding gesture displayed in the background. The results indicate that the \mbox{T-Att} module is able to focus on the key frames while ignores the preparation and end frames.




\subsection{Evaluation of Hyper-Parameters}

The choice of hyper-parameters is important to the final recognition accuracy for the network. This subsection investigates four hyper-parameters: the multi-head number, the dimension of each head, weight sharing for \mbox{J-Att} and \mbox{T-Att}, and the dropout rate.

\textbf{Multi-head number}. Multi-head attention is used to learn more attention information from different representation subspaces at different positions. Table~\ref{tab:mutihead&head-dimension&dropout} lists the results with different numbers of heads. For a fair comparison, the dimensions of key, query and value vectors in each head are adjusted to ensure the network shares the same parameter size. When the multi-head number increases from 1 to 8, the performance improves gradually. When the multi-head number is greater than 4, the recognition accuracy tends to be saturated. The best performance is achieved with 8 heads.

\textbf{Dimension of each head}. To investigate the choice for the dimensions of key, query and value vectors in each head, experiments with different dimensions are conducted. The results are illustrated in Table~\ref{tab:mutihead&head-dimension&dropout}. The small dimension has weak feature extraction abilities, while the oversize dimension could cause overfitting. The optimal dimension is 32.

\textbf{Dropout rate}. Since the three benchmark datasets are small in size (less than 2800 samples), the network is prone to overfitting. The dropout operation in attention module can help alleviate overfitting. Results with different dropout rates are listed in Table~\ref{tab:mutihead&head-dimension&dropout}. The best dropout rate is 0.1, excessive dropout rate will lead to underfitting.

\textbf{Weight sharing}. Results with different weight sharing strategies are given in Table~\ref{tab:weight-sharing}. The best performance is achieved with weight sharing on both \mbox{J-Att} and \mbox{T-Att} modules. Weight sharing helps to reduce the network parameters, which contributes to avoiding overfitting.

\begin{table}[!t]
	\centering
	\caption{Results with different number of heads, head dimensions, and dropout rates on SHREC'17 Track dataset with 14 gestures setting.}
	\label{tab:mutihead&head-dimension&dropout}
	\begin{tabular}{c|c||c|c||c|c}
		\toprule
		\textbf{\makecell[l]{Num of\\ Head}} & \textbf{\makecell[l]{Acc.\\(\%)}}& \textbf{\makecell[l]{Head\\Dimension}} & \textbf{\makecell[l]{Acc.\\(\%)}} & \textbf{\makecell[l]{Dropout\\Rate}} & \textbf{\makecell[l]{Acc.\\(\%)}}\\	
		\hline
		1 & 92.14 & 8 & 94.64 & 0 & 94.64\\
		2 & 93.93 & 16 & 94.88& \textbf{0.1} & \textbf{95.00}\\
		4 & 94.64 & \textbf{32} & \textbf{95.00} & 0.2 & 94.64\\
		6 & 94.88 & 64 & 94.76 & 0.3 & 93.21\\
		\textbf{8} & \textbf{95.00} & 128 & 94.52 & 0.4 & 92.02\\
		\bottomrule
	\end{tabular}
\end{table}

\begin{table}[!t]
	\centering
	\caption{Results of different weight sharing strategies for \mbox{J-Att} and \mbox{T-Att} on SHREC'17 Track dataset with 14 gestures setting.}
	\label{tab:weight-sharing}
	\begin{tabular}{c|cc|c} 
		\toprule
		\textbf{Weight Sharing} & \textbf{\mbox{J-Att}} & \textbf{\mbox{T-Att}} & \textbf{Acc. (\%)}  \\ 
		\hline
		\multirow{3}{*}{Ablations} & \xmark & \xmark &  94.52 \\ 
		\cline{2-4}
		& \xmark & \cmark &  94.76 \\ 
		\cline{2-4}
		& \cmark & \xmark & 94.88  \\ 
		\hline
		\textbf{HAN} & \cmark & \cmark & \textbf{95.00}  \\
		\bottomrule
	\end{tabular}
\end{table}


\section{Conclusion}
\label{sec5}


This paper proposes an efficient hierarchical self-attention network (HAN) for skeleton-based gesture recognition. Four self-attention modules are designed to capture spatial and temporal features for skeleton sequence gradually. For the skeleton at a specific frame, the skeleton joints are divided into 6 parts (5 fingers and the palm). Our method applies the joint self-attention module (\mbox{J-Att}) to extract the features of each finger. Then the finger self-attention module (\mbox{F-Att}) is designed to aggregate features of the whole hand. In terms of temporal features, the temporal self-attention module (\mbox{T-Att}) is utilized to capture the temporal dynamics of the fingers and the entire hand. Finally, these features are fused by the fusion self-attention module (\mbox{Fusion-Att}) for gesture classification. Experiments show that our method achieves competitive results on three gesture recognition datasets with much lower computational complexity.

\ifCLASSOPTIONcaptionsoff
\newpage
\fi



\bibliographystyle{IEEEtran}
\bibliography{IEEEabrv,ref}

\begin{thebibliography}{10}
\providecommand{\url}[1]{#1}
\csname url@samestyle\endcsname
\providecommand{\newblock}{\relax}
\providecommand{\bibinfo}[2]{#2}
\providecommand{\BIBentrySTDinterwordspacing}{\spaceskip=0pt\relax}
\providecommand{\BIBentryALTinterwordstretchfactor}{4}
\providecommand{\BIBentryALTinterwordspacing}{\spaceskip=\fontdimen2\font plus
\BIBentryALTinterwordstretchfactor\fontdimen3\font minus
  \fontdimen4\font\relax}
\providecommand{\BIBforeignlanguage}[2]{{%
\expandafter\ifx\csname l@#1\endcsname\relax
\typeout{** WARNING: IEEEtran.bst: No hyphenation pattern has been}%
\typeout{** loaded for the language `#1'. Using the pattern for}%
\typeout{** the default language instead.}%
\else
\language=\csname l@#1\endcsname
\fi
#2}}
\providecommand{\BIBdecl}{\relax}
\BIBdecl

\bibitem{chen2017motion}
X.~Chen, H.~Guo, G.~Wang, and L.~Zhang, ``Motion feature augmented recurrent
  neural network for skeleton-based dynamic hand gesture recognition,'' in
  \emph{Proc. Int. Conf. Image Process. (ICIP)}, Beijing, China, Sep. 2017, pp.
  2881--2885.

\bibitem{hou2018spatial}
J.~Hou, G.~Wang, X.~Chen, J.-H. Xue, R.~Zhu, and H.~Yang, ``Spatial-temporal
  attention res-tcn for skeleton-based dynamic hand gesture recognition,'' in
  \emph{Proc. Eur. Conf. Comput. Vis. Workshops (ECCVW)}, Munich, Germany, Sep.
  2018, pp. 273--286.

\bibitem{nunez2018convolutional}
J.~C. Nunez, R.~Cabido, J.~J. Pantrigo, A.~S. Montemayor, and J.~F. Velez,
  ``Convolutional neural networks and long short-term memory for skeleton-based
  human activity and hand gesture recognition,'' \emph{Pattern Recognit.},
  vol.~76, pp. 80--94, Apr. 2018.

\bibitem{yan2018spatial}
S.~Yan, Y.~Xiong, and D.~Lin, ``Spatial temporal graph convolutional networks
  for skeleton-based action recognition,'' in \emph{Proc. AAAI Conf. Artif.
  Intell. (AAAI)}, New Orleans, LA, United states, Feb. 2018, pp. 7444--7452.

\bibitem{liu2020decoupled}
J.~Liu, Y.~Liu, Y.~Wang, V.~Prinet, S.~Xiang, and C.~Pan, ``Decoupled
  representation learning for skeleton-based gesture recognition,'' in
  \emph{Proc. IEEE Comput. Soc. Conf. Comput. Vis. Pattern Recognit. (CVPR)},
  Virtual, Online, United states, Jun. 2020, pp. 5751--5760.

\bibitem{de2016skeleton}
Q.~De~Smedt, H.~Wannous, and J.-P. Vandeborre, ``Skeleton-based dynamic hand
  gesture recognition,'' in \emph{Proc. IEEE Comput. Soc. Conf. Comput. Vis.
  Pattern Recognit. Workshops (CVPRW)}, Las Vegas, NV, United states, Jun.
  2016, pp. 1206 -- 1214.

\bibitem{weng2018deformable}
J.~Weng, M.~Liu, X.~Jiang, and J.~Yuan, ``Deformable pose traversal convolution
  for 3d action and gesture recognition,'' in \emph{Proc. Eur. Conf. Comput.
  Vis. (ECCV)}, Munich, Germany, Sep. 2018, pp. 142--157.

\bibitem{devineau2018deep}
G.~Devineau, F.~Moutarde, W.~Xi, and J.~Yang, ``Deep learning for hand gesture
  recognition on skeletal data,'' in \emph{Proc. IEEE Int. Conf. Autom. Face
  Gesture Recognit. (FG)}, Xi'an, China, May 2018, pp. 106--113.

\bibitem{chen2019construct}
Y.~Chen, L.~Zhao, X.~Peng, J.~Yuan, and D.~N. Metaxas, ``Construct dynamic
  graphs for hand gesture recognition via spatial-temporal attention,'' in
  \emph{Proc. Brit. Mach. Vis. Conf. (BMVC)}, Cardiff, United kingdom, Sep.
  2019.

\bibitem{nguyen2019neural}
X.~S. Nguyen, L.~Brun, O.~L{\'e}zoray, and S.~Bougleux, ``A neural network
  based on spd manifold learning for skeleton-based hand gesture recognition,''
  in \emph{Proc. IEEE Comput. Soc. Conf. Comput. Vis. Pattern Recognit.
  (CVPR)}, Long Beach, CA, United states, Jun. 2019, pp. 12\,028--12\,037.

\bibitem{fan2016action}
J.~Fan, Z.~Zha, and X.~Tian, ``Action recognition with novel high-level pose
  features,'' in \emph{Proc. IEEE Int. Conf. Multimedia Expo Workshops
  (ICMEW)}, Seattle, WA, United states, Jul. 2016, pp. 1--6.

\bibitem{ohn2013joint}
E.~Ohn-Bar and M.~Trivedi, ``Joint angles similarities and hog2 for action
  recognition,'' in \emph{Proc. IEEE Comput. Soc. Conf. Comput. Vis. Pattern
  Recognit. Workshops (CVPRW)}, Portland, OR, United states, Jun. 2013, pp.
  465--470.

\bibitem{devanne20143}
M.~Devanne, H.~Wannous, S.~Berretti, P.~Pala, M.~Daoudi, and A.~Del~Bimbo,
  ``3-d human action recognition by shape analysis of motion trajectories on
  riemannian manifold,'' \emph{IEEE Trans. Cybern.}, vol.~45, no.~7, pp.
  1340--1352, Jul. 2015.

\bibitem{evangelidis2014skeletal}
G.~Evangelidis, G.~Singh, and R.~Horaud, ``Skeletal quads: Human action
  recognition using joint quadruples,'' in \emph{Proc. Int. Conf. Pattern
  Recognit. (ICPR)}, Stockholm, Sweden, Aug. 2014, pp. 4513--4518.

\bibitem{vemulapalli2014human}
R.~Vemulapalli, F.~Arrate, and R.~Chellappa, ``Human action recognition by
  representing 3d skeletons as points in a lie group,'' in \emph{Proc. IEEE
  Comput. Soc. Conf. Comput. Vis. Pattern Recognit. (CVPR)}, Columbus, OH,
  United states, Jun. 2014, pp. 588--595.

\bibitem{du2015hierarchical}
Y.~Du, W.~Wang, and L.~Wang, ``Hierarchical recurrent neural network for
  skeleton based action recognition,'' in \emph{Proc. IEEE Comput. Soc. Conf.
  Comput. Vis. Pattern Recognit. (CVPR)}, Boston, MA, United states, Jun. 2015,
  pp. 1110--1118.

\bibitem{song2017end}
S.~Song, C.~Lan, J.~Xing, W.~Zeng, and J.~Liu, ``An end-to-end spatio-temporal
  attention model for human action recognition from skeleton data,'' in
  \emph{Proc. AAAI Conf. Artif. Intell. (AAAI)}, San Francisco, CA, United
  states, Feb. 2017, pp. 4263--4270.

\bibitem{zhang2017view}
P.~Zhang, C.~Lan, J.~Xing, W.~Zeng, J.~Xue, and N.~Zheng, ``View adaptive
  recurrent neural networks for high performance human action recognition from
  skeleton data,'' in \emph{Proc. IEEE Int. Conf. Comput. Vis. (ICCV)}, Venice,
  Italy, Oct. 2017, pp. 2136--2145.

\bibitem{li2018independently}
S.~Li, W.~Li, C.~Cook, C.~Zhu, and Y.~Gao, ``Independently recurrent neural
  network (indrnn): Building a longer and deeper rnn,'' in \emph{Proc. IEEE
  Comput. Soc. Conf. Comput. Vis. Pattern Recognit. (CVPR)}, Salt Lake City,
  UT, United states, Jun. 2018, pp. 5457--5466.

\bibitem{si2018skeleton}
C.~Si, Y.~Jing, W.~Wang, L.~Wang, and T.~Tan, ``Skeleton-based action
  recognition with spatial reasoning and temporal stack learning,'' in
  \emph{Proc. Eur. Conf. Comput. Vis. (ECCV)}, Munich, Germany, Sep. 2018, pp.
  103--118.

\bibitem{wang2017modeling}
H.~Wang and L.~Wang, ``Modeling temporal dynamics and spatial configurations of
  actions using two-stream recurrent neural networks,'' in \emph{Proc. IEEE
  Comput. Soc. Conf. Comput. Vis. Pattern Recognit. (CVPR)}, Honolulu, HI,
  United states, Jul. 2017, pp. 3633--3642.

\bibitem{kim2017interpretable}
T.~S. Kim and A.~Reiter, ``Interpretable 3d human action analysis with temporal
  convolutional networks,'' in \emph{Proc. IEEE Comput. Soc. Conf. Comput. Vis.
  Pattern Recognit. Workshops (CVPRW)}, Honolulu, HI, United states, Jul. 2017,
  pp. 1623--1631.

\bibitem{cao2018skeleton}
C.~Cao, C.~Lan, Y.~Zhang, W.~Zeng, H.~Lu, and Y.~Zhang, ``Skeleton-based action
  recognition with gated convolutional neural networks,'' \emph{{IEEE} Trans.
  Circuits Syst. Video Technol.}, vol.~29, no.~11, pp. 3247--3257, Nov. 2019.

\bibitem{banerjee2020fuzzy}
A.~Banerjee, P.~K. Singh, and R.~Sarkar, ``Fuzzy integral based cnn classifier
  fusion for 3d skeleton action recognition,'' \emph{{IEEE} Trans. Circuits
  Syst. Video Technol.}, Aug. 2020.

\bibitem{liu2017enhanced}
M.~Liu, H.~Liu, and C.~Chen, ``Enhanced skeleton visualization for view
  invariant human action recognition,'' \emph{Pattern Recognit.}, vol.~68, pp.
  346--362, Aug. 2017.

\bibitem{shi2019two}
L.~Shi, Y.~Zhang, J.~Cheng, and H.~Lu, ``Two-stream adaptive graph
  convolutional networks for skeleton-based action recognition,'' in
  \emph{Proc. IEEE Comput. Soc. Conf. Comput. Vis. Pattern Recognit. (CVPR)},
  Long Beach, CA, United states, Jun. 2019, pp. 12\,026--12\,035.

\bibitem{li2019actional}
M.~Li, S.~Chen, X.~Chen, Y.~Zhang, Y.~Wang, and Q.~Tian, ``Actional-structural
  graph convolutional networks for skeleton-based action recognition,'' in
  \emph{Proc. IEEE Comput. Soc. Conf. Comput. Vis. Pattern Recognit. (CVPR)},
  Long Beach, CA, United states, Jun. 2019, pp. 3595--3603.

\bibitem{kong2021symmetrical}
J.~Kong, H.~Deng, and M.~Jiang, ``Symmetrical enhanced fusion network for
  skeleton-based action recognition,'' \emph{{IEEE} Trans. Circuits Syst. Video
  Technol.}, Jan. 2021.

\bibitem{si2019attention}
C.~Si, W.~Chen, W.~Wang, L.~Wang, and T.~Tan, ``An attention enhanced graph
  convolutional lstm network for skeleton-based action recognition,'' in
  \emph{Proc. IEEE Comput. Soc. Conf. Comput. Vis. Pattern Recognit. (CVPR)},
  Long Beach, CA, United states, Jun. 2019, pp. 1227--1236.

\bibitem{vaswani2017attention}
A.~Vaswani, N.~Shazeer, N.~Parmar, J.~Uszkoreit, L.~Jones, A.~N. Gomez,
  {\L}.~Kaiser, and I.~Polosukhin, ``Attention is all you need,'' in
  \emph{Proc. Adv. neural inf. proces. syst. (NeurIPS)}, Long Beach, CA, United
  states, Dec. 2017, pp. 5999--6009.

\bibitem{dosovitskiy2020image}
A.~Dosovitskiy, L.~Beyer, A.~Kolesnikov, D.~Weissenborn, X.~Zhai,
  T.~Unterthiner, M.~Dehghani, M.~Minderer, G.~Heigold, S.~Gelly, J.~Uszkoreit,
  and N.~Houlsby, ``An image is worth 16x16 words: Transformers for image
  recognition at scale,'' in \emph{Proc. Int. Conf. Learn. Representations
  (ICLR)}, 2021.

\bibitem{carion2020end}
N.~Carion, F.~Massa, G.~Synnaeve, N.~Usunier, A.~Kirillov, and S.~Zagoruyko,
  ``End-to-end object detection with transformers,'' in \emph{Proc. Eur. Conf.
  Comput. Vis. (ECCV)}, Glasgow, United kingdom, Aug. 2020, pp. 213--229.

\bibitem{zheng2020rethinking}
S.~Zheng, J.~Lu, H.~Zhao, X.~Zhu, Z.~Luo, Y.~Wang, Y.~Fu, J.~Feng, T.~Xiang,
  P.~H. Torr \emph{et~al.}, ``Rethinking semantic segmentation from a
  sequence-to-sequence perspective with transformers,'' \emph{arXiv preprint
  arXiv:2012.15840}, 2020.

\bibitem{de2017shrec}
Q.~De~Smedt, H.~Wannous, J.-P. Vandeborre, J.~Guerry, B.~Le~Saux, and
  D.~Filliat, ``Shrec'17 track: 3d hand gesture recognition using a depth and
  skeletal dataset,'' in \emph{Proc. Eurographics Workshop on 3D Object Retr.
  (3DOR)}, Lyon, France, Apr. 2017, pp. 33--38.

\bibitem{garcia2018first}
G.~Garcia-Hernando, S.~Yuan, S.~Baek, and T.-K. Kim, ``First-person hand action
  benchmark with rgb-d videos and 3d hand pose annotations,'' in \emph{Proc.
  IEEE Comput. Soc. Conf. Comput. Vis. Pattern Recognit. (CVPR)}, Salt Lake
  City, UT, United states, Jun. 2018, pp. 409--419.

\bibitem{de2019heterogeneous}
Q.~De~Smedt, H.~Wannous, and J.-P. Vandeborre, ``Heterogeneous hand gesture
  recognition using 3d dynamic skeletal data,'' \emph{Comput. Vis. Image
  Underst.}, vol. 181, pp. 60--72, Apr. 2019.

\bibitem{tu2018skeleton}
J.~Tu, M.~Liu, and H.~Liu, ``Skeleton-based human action recognition using
  spatial temporal 3d convolutional neural networks,'' in \emph{Proc. IEEE Int.
  Conf. Multimedia Expo (ICME)}, San Diego, CA, United states, Jul. 2018, pp.
  1--6.

\bibitem{liu2018learning}
H.~Liu, J.~Tu, M.~Liu, and R.~Ding, ``Learning explicit shape and motion
  evolution maps for skeleton-based human action recognition,'' in \emph{Proc.
  IEEE Int. Conf. Acoust. Speech Signal Process. (ICASSP)}, Calgary, AB,
  Canada, Apr. 2018, pp. 1333--1337.

\bibitem{oreifej2013hon4d}
O.~Oreifej and Z.~Liu, ``Hon4d: Histogram of oriented 4d normals for activity
  recognition from depth sequences,'' in \emph{Proc. IEEE Comput. Soc. Conf.
  Comput. Vis. Pattern Recognit. (CVPR)}, Portland, OR, United states, Jun.
  2013, pp. 716--723.

\bibitem{de2017dynamic}
Q.~De~Smedt, ``Dynamic hand gesture recognition-from traditional handcrafted to
  recent deep learning approaches,'' Ph.D. dissertation, Universit{\'e} de
  Lille 1, Sciences et Technologies; CRIStAL UMR 9189, 2017.

\bibitem{caputo2018comparing}
F.~M. Caputo, P.~Prebianca, A.~Carcangiu, L.~D. Spano, and A.~Giachetti,
  ``Comparing 3d trajectories for simple mid-air gesture recognition,''
  \emph{Comput. Graph.}, vol.~73, pp. 17--25, Jun. 2018.

\bibitem{boulahia2017dynamic}
S.~Y. Boulahia, E.~Anquetil, F.~Multon, and R.~Kulpa, ``Dynamic hand gesture
  recognition based on 3d pattern assembled trajectories,'' in \emph{Proc. Int.
  Conf. Image Process. Theory (IPTA)}, Montreal, QC, Canada, Nov. 2017, pp.
  1--6.

\bibitem{zanfir2013moving}
M.~Zanfir, M.~Leordeanu, and C.~Sminchisescu, ``The moving pose: An efficient
  3d kinematics descriptor for low-latency action recognition and detection,''
  in \emph{Proc. IEEE Int. Conf. Comput. Vis. (ICCV)}, Sydney, NSW, Australia,
  Dec. 2013, pp. 2752--2759.

\bibitem{zhang2016efficient}
X.~Zhang, Y.~Wang, M.~Gou, M.~Sznaier, and O.~Camps, ``Efficient temporal
  sequence comparison and classification using gram matrix embeddings on a
  riemannian manifold,'' in \emph{Proc. IEEE Comput. Soc. Conf. Comput. Vis.
  Pattern Recognit. (CVPR)}, Las Vegas, NV, United states, Jun. 2016, pp.
  4498--4507.

\bibitem{garcia2017transition}
G.~Garcia-Hernando and T.-K. Kim, ``Transition forests: Learning discriminative
  temporal transitions for action recognition and detection,'' in \emph{Proc.
  IEEE Comput. Soc. Conf. Comput. Vis. Pattern Recognit. (CVPR)}, Honolulu, HI,
  United states, Jul. 2017, pp. 432--440.

\bibitem{huang2017riemannian}
Z.~Huang and L.~Van~Gool, ``A riemannian network for spd matrix learning,'' in
  \emph{Proc. AAAI Conf. Artif. Intell. (AAAI)}, San Francisco, CA, United
  states, Feb. 2017, pp. 2036--2042.

\bibitem{huang2018building}
Z.~Huang, J.~Wu, and L.~Van~Gool, ``Building deep networks on grassmann
  manifolds,'' in \emph{Proc. AAAI Conf. Artif. Intell. (AAAI)}, New Orleans,
  LA, United states, Feb. 2018, pp. 3279--3286.

\end{thebibliography}

\vfill


\end{document}